\documentclass{article}

\usepackage[final,dblblindworkshop]{neurips_2026}

\workshoptitle{CODEC-FM}

\usepackage[utf8]{inputenc}
\usepackage[T1]{fontenc}
\usepackage{microtype}
\usepackage{graphicx}
\usepackage{float}
\usepackage{booktabs}
\usepackage{tabularx}
\usepackage{adjustbox}
\usepackage{makecell}
\usepackage{multirow}
\usepackage{xcolor}
\usepackage{hyperref}
\usepackage{url}
\usepackage{pgfplots}
\usepackage{amsmath}
\usepackage{amssymb}
\usepgfplotslibrary{groupplots}
\usetikzlibrary{arrows.meta,backgrounds,calc,fit,positioning}
\pgfplotsset{compat=1.18}

\definecolor{wandbdivine}{HTML}{79C9BA}
\definecolor{armA}{HTML}{9CA3AF}
\definecolor{armB}{HTML}{3B82F6}
\definecolor{armC}{HTML}{14B8A6}
\definecolor{armD}{HTML}{D946EF}
\definecolor{armE}{HTML}{F59E0B}
\definecolor{armF}{HTML}{8B5CF6}
\definecolor{nanortblue}{HTML}{155B8A}
\definecolor{controlorange}{HTML}{F5A623}
\definecolor{methodpurple}{HTML}{D946EF}
\definecolor{targetgray}{HTML}{6B7280}
\definecolor{controlgray}{HTML}{9CA3AF}
\definecolor{targetgreen}{HTML}{2E7D32}
\definecolor{archink}{HTML}{172033}
\definecolor{archmuted}{HTML}{667085}
\definecolor{archline}{HTML}{98A2B3}
\definecolor{archblue}{HTML}{2563EB}
\definecolor{archbluefill}{HTML}{EFF6FF}
\definecolor{archgreen}{HTML}{15803D}
\definecolor{archgreenfill}{HTML}{F0FDF4}
\definecolor{archpurple}{HTML}{C026D3}
\definecolor{archpurplefill}{HTML}{FDF4FF}
\definecolor{archamber}{HTML}{B45309}
\definecolor{archamberfill}{HTML}{FFF7ED}
\definecolor{archgrayfill}{HTML}{F8FAFC}
\definecolor{archrose}{HTML}{C98282}
\definecolor{archrosefill}{HTML}{F6DEDE}
\definecolor{archyellow}{HTML}{D39A2C}
\definecolor{archyellowfill}{HTML}{F6D88A}
\definecolor{archblockblue}{HTML}{7890B2}
\definecolor{archblockbluefill}{HTML}{CCD8E8}
\definecolor{archorange}{HTML}{D97A00}
\definecolor{archburgundy}{HTML}{7A1F3D}
\definecolor{archburgundyfill}{HTML}{F6E8ED}
\title{Speedrunning Relational Foundation Model Pretraining}

  \usetikzlibrary{arrows.meta,calc,positioning}
  \definecolor{archslate}{HTML}{6E86A8}
  \definecolor{archslatefill}{HTML}{E3E9F2}
  \definecolor{archshell}{HTML}{F3F5F8}

\author{
\textbf{Gabriel Wedson Mendonça de Souza Veloso}$^{1,2}$
\quad
\textbf{Luan Borges Teodoro Reis Sena}$^{1,2}$\\
\textbf{Denis Oliveira Correa}$^{1}$
\quad
\textbf{Francisco Galuppo Azevedo}$^{1,2}$\\[0.4em]
$^{1}$Kunumi Institute, Belo Horizonte, Minas Gerais, Brazil\\
$^{2}$Universidade Federal de Minas Gerais, Belo Horizonte, Minas Gerais, Brazil\\
\texttt{\{gabriel.veloso, luan.borges, denis, francisco\}@kunumi.com}
}

\begin{document}

\maketitle

\begin{abstract}
Progress on foundation model pretraining is gated by who can afford a trial, and relational pretraining is no exception. Recipes are developed inside individual labs, reported under private protocols, and do not accumulate. Speedrun leaderboards remove that barrier by decentralizing the search rather than the training run. The problem is fixed, the stack is open, entries are ranked by wall-clock time to a quality target, and one GPU is enough to hold the record. We introduce modded-nanoRT, the first such leaderboard for Relational Transformers. Which database to hold out decides whether the benchmark can resolve improvements at all, so we fix it by calibration, ranking six candidates by how sharply a 45-minute learning curve separates improvement from seed noise. We open the leaderboard with typed trilinear attention, a relational operator that conditions attention scores and values on the foreign key traversed, together with Muon and SIGReg. The recipe reaches the target in $305.63$ seconds on one NVIDIA H200, $8.11\times$ faster than the reference implementation, and wins 13 of 15 pairs outside the development database. We will release the benchmark, its frozen artifacts, and the record runs. Our code is available at \url{https://github.com/kunumi/nanoRT}.
\end{abstract}

\section{Introduction}

Most machine learning data sits in relational databases, split across tables linked by primary--foreign key relationships. Standard pipelines flatten those links into a task-specific table, so data preparation repeats for every new database and prediction problem. Relational foundation models learn from the schema itself. RelBench standardized this setting across databases and predictive tasks \citep{robinson2024relbench}, and the Relational Transformer (RT) showed that one schema-aware model applies to unseen tasks without task-specific parameters or in-context examples \citep{ranjan2026relational}.

An architectural, optimizer, or regularization change takes hours to evaluate even at small scale, and the labs that can run such trials at volume are the labs with clusters, so recipe knowledge accumulates where the compute is and everyone else is priced out of the loop that produces it. Community speedruns answer this by pooling a different resource. Fixing the model family, the data, the hardware, and the target, and ranking submissions by time to reach it, turned recipe search for language model pretraining into a leaderboard that dozens of contributors each advanced by seconds \citep{modded_nanogpt_2024}; modded-nanoTabPFN did the same for tabular models \citep{pfefferle2025nanotabpfn, ozturk2026speedrunning}. What is decentralized here is not the training run but the search over it. Contributors hold one GPU each, work independently on disjoint parts of the design space, and their findings compose because every entry is measured on the same clock.

No such artifact exists for relational foundation models, and the format does not transfer mechanically. Language and tabular speedruns train on one fixed corpus. Relational pretraining mixes databases with different schemas, and the quantity of interest is zero-shot transfer to a database the model has never seen, so the mixture and an entire held-out database must be fixed in advance.

We introduce \emph{modded-nanoRT}, an open speedrun for pretraining a nano-scale Relational Transformer on a single GPU and evaluating it zero-shot on a held-out database. Table~\ref{tab:protocol} lists what is fixed; the training stack is open, and submissions are ranked by wall-clock time to the target. Choosing the held-out database is the design problem the relational setting adds, and we settle it by calibration. All six candidates run for 45 minutes over multiple seeds, ranked by how sharply their learning curves separate improvements from seed noise.

We open the leaderboard with three changes to the RT recipe. We introduce \emph{typed trilinear attention}. RT uses the schema only as an attention mask, so every foreign key shares one compatibility function; we make the relation a third multiplicative factor in the attention score and the message, so neighbors reached by different foreign keys are no longer scored alike. We add SIGReg as a representation-level regularizer over relational columns \citep{balestriero2025lejepa}, following tabular work that reports faster convergence under it \citep{merchant2026optimizing}. Muon orthogonalizes the matrix-valued updates of the backbone, carried over from the language model speedruns \citep{jordan2024muon}.

The record reaches the target in \textbf{305.63 seconds} on one NVIDIA H200, $8.11\times$ faster than the reference implementation. Matched three-seed comparisons attribute the gain to Muon, and to typed attention only in combination with it; leave-one-database-out runs win 13 of 15 pairs outside development. Verification consumes $59.9\%$ of the record, so the next entry should look at the evaluation path.
\section{Related Work}
\label{sec:related}

\paragraph{Relational foundation models.} Relational deep learning has standardized benchmarks \citep{robinson2024relbench} and pretrained schema-aware models \citep{ranjan2026relational, ranjan2026largescale}, and synthetic multi-table generation addresses the scarcity of relational pretraining data \citep{kothapalli2026plurel}. These fix what to train on. None fix a protocol under which training-efficiency results can be compared, so recipe improvements are reported against private baselines and do not compose.

\paragraph{Time-to-quality benchmarks.} DAWNBench introduced time-to-accuracy as a joint criterion for algorithmic and systems improvements \citep{coleman2018dawnbench}, MLPerf Training formalized quality targets under stochastic time to solution \citep{mattson2020mlperf}, and Cramming audits a language model pipeline under a one-GPU, one-day budget \citep{geiping2022cramming}. Speedrun leaderboards inherit this criterion and add an open submission process, so contributors accumulate improvements between benchmark rounds. modded-nanoRT carries that to relational pretraining, where the database mixture and the held-out database create their own validity constraints.

\paragraph{Typed and higher-order attention.} Typed edges are handled elsewhere by relation-specific message transformations \citep{schlichtkrull2017rgcn}, and trilinear scores appear in the 2-simplicial Transformer \citep{clift2019logic, roy2025fast}. Our third factor is the database relation on a query--neighbor edge rather than a second neighbor, so typed trilinear attention is relation-conditioned message passing rather than a simplicial architecture.
\section{The modded-nanoRT challenge}
\label{sec:challenge}

A full run fits on one accelerator. Table~\ref{tab:protocol} lists the fixed quantities; the architecture, optimizer, regularization, context sampling algorithm, precision, and systems implementation remain open.

\begin{table}[h]
\centering
\small
\begin{tabular}{ll}
\toprule
Pretraining tasks & RelBench v1 binary classification tasks, minus rel-Avito and rel-Event \\
Mixture & temperature sampling, $\alpha=0.5$ \\
Held-out database & rel-Avito \\
Metric & pooled AUROC on rel-Avito \\
Target & $0.81621$ \\
Reference model & RT, 12 blocks, width 256, 8 heads \\
Trainable parameters & 19.3M \\
Hardware & 1$\times$ NVIDIA H200 \\
Clock & start to first evaluation reaching target \\
\bottomrule
\end{tabular}
\caption{The modded-nanoRT protocol.}
\label{tab:protocol}
\end{table}

\paragraph{Training mixture.} RelBench v1 provides seven databases. We keep only binary classification tasks, so that one metric applies across every database, and exclude rel-Event for confirmed temporal leakage. Six remain as candidates, with seed counts differing by orders of magnitude: rel-Amazon, rel-HM, rel-Stack, rel-Avito, rel-Trial, and rel-F1. One of the six is held out and excluded from the mixture, which therefore spans five. Sampling in proportion to those counts lets the largest database dominate the gradient and turns multi-database pretraining into single-database training; sampling uniformly over-weights the small schemas. We therefore temper the mixture, in the convention used for multilingual pretraining \citep{xue2021mt5}, $w(d)\propto n(d)^{\alpha}$, $\alpha=0.5$, where $n(d)$ is the seed count of database $d$ and the weights are normalized to sum to one, halfway between proportional ($\alpha=1$) and uniform ($\alpha=0$). Tempering is applied when the training artifact is built, by subsampling each database's seed list until its share of the seeds equals $w(d)$. The mixture is frozen as part of the protocol, because changing it changes what the metric measures.

\paragraph{Held-out database.} Evaluation is zero-shot on a database excluded from the mixture, and which database plays that role decides whether the benchmark can resolve improvements at all. A database that saturates early, stalls at the noise floor, or is dominated by seed variance cannot rank entries. We settled it empirically, running a 45-minute leave-one-out calibration for each of six candidates over multiple seeds and ranking them by
\begin{equation}
\mathrm{score}(d)\;=\;\frac{\Delta\mathrm{AUROC}(d)\cdot s(d)}{\sigma(d)},
\label{eq:holdout-score}
\end{equation}
where $\Delta\mathrm{AUROC}$ is the gain from initialization to 45 minutes, $s$ is the regression slope of the AUROC curve over the final 20\% of wall-clock time, and $\sigma$ is the standard deviation across seeds. rel-Avito ranked first. The target, $0.81621$, is the best pooled AUROC its calibration run reached inside the 45-minute budget, at $41$:$20$ of $45$:$23$ (Appendix~\ref{app:calibration}); the validation artifact was frozen alongside it with seeds capped at $50\,000$ and a fixed hop radius and text embeddings. Records are therefore speedups over the reference implementation, resolved to the width of each run's evaluation grid.

\paragraph{Evaluation.} We evaluate on the frozen validation artifact for the two rel-Avito tasks, \texttt{user-visits} and \texttt{user-clicks} \citep{robinson2024relbench}. At each scheduled evaluation we concatenate predictions and labels across both tasks and compute one AUROC. Pooling weights examples rather than tasks and confounds within-task ranking with cross-task score alignment. Only the pooled AUROC determines record eligibility.

\paragraph{Reference baseline.} The baseline is the Relational Transformer of \citet{ranjan2026relational}, reduced to nano scale. RT builds a temporally valid context around each seed row, masks the target cell, and represents every retained cell by its value, datatype, and frozen table--column text embedding; column, feature/foreign-key, and full attention mix these tokens before a datatype-specific decoder predicts the masked value. The baseline uses 12 blocks, width 256, eight heads, and a gated SiLU MLP of width 1024, for 19.3M trainable parameters. Pretraining uses bfloat16, contexts of 1024 cells, batches of 256, and AdamW at learning rate $10^{-3}$, weight decay $0.1$, with linear warmup over the first 20\% of updates and linear decay after.

\paragraph{Rules.} modded-nanoRT holds the problem fixed \citep{mattson2020mlperf,modded_nanogpt_2024,ozturk2026speedrunning}. A submission must use one H200, the prescribed training artifact, the frozen rel-Avito validation artifact, and the official metric, target, evaluation schedule, and timing boundary. Weights start from a declared random seed; a submission may not load a checkpoint, train on rel-Avito, reveal masked targets, or modify the validation pipeline. The budget is 45 minutes, and runs that miss the target stay in the analysis as right-censored failures.
\section{The record recipe}
\label{sec:recipe}

\paragraph{Typed trilinear attention.} Standard RT scores a query cell $i$ and a neighboring cell $j$ bilinearly, so all foreign-key types share the same compatibility function and a relation enters the model only as an attention mask. Two neighbors carrying the same value are then scored identically whether they were reached through \texttt{review}$\to$\texttt{product} or \texttt{review}$\to$\texttt{customer}, and the model must recover the difference from cell content alone, separately in every context it samples. We attach the relation type $\rho(i,j)$ as a third factor in both the attention score and the message, driven by a frozen text embedding of the foreign-key description.

Relation-specific message transformations are standard \citep{schlichtkrull2017rgcn}, but they allocate parameters per relation type and so cannot type an edge in a schema the model has never seen; the embedding comes from an external text encoder, so a new schema adds no parameters. Appendix~\ref{app:fk-ablation} shows that the distinctness across relation types is what matters, not its content.

The factor is multiplicative rather than additive, following the trilinear principle of higher-order attention \citep{clift2019logic,roy2025fast}. An additive term shifts the scores of an edge class uniformly and gives the relation the status of a query-independent prior over edge types; multiplying inside the head dimension lets the relation select which coordinates of the query--key match are counted, so one query can rank the same neighbor differently depending on how it was reached.

Let $x_i$ and $r_\rho$ denote the RMS-normalized cell and relation inputs to this sublayer.
\begin{equation}
\begin{aligned}
q_i^h&=\mathrm{RMSNorm}(W_q^h x_i), &
k_j^h&=\mathrm{RMSNorm}(W_k^h x_j),\\
a_\rho^h&=\mathrm{RMSNorm}(W_{kr}^h r_\rho), &
v_j^h&=W_v^h x_j,\qquad u_\rho^h=W_{vr}^h r_\rho .
\end{aligned}
\label{eq:trilinear-factors}
\end{equation}
The typed score, edge-normalized attention, and message are
\begin{equation}
e_{ij}^h
=\frac{1}{\sqrt{d_h}}
\sum_{c=1}^{d_h}q_{i,c}^h k_{j,c}^h a_{\rho(i,j),c}^h,
\qquad
\alpha_{ij}^h
=\frac{\exp(e_{ij}^h)}
{\sum_{\ell\in\mathcal N(i)}\exp(e_{i\ell}^h)},
\label{eq:trilinear-score}
\end{equation}
\begin{equation}
o_i^h=\sum_{j\in\mathcal N(i)}
\alpha_{ij}^h\left(v_j^h\odot u_{\rho(i,j)}^h\right),
\qquad
x_i\leftarrow x_i+\gamma\odot W_o
\left[o_i^1;\ldots;o_i^H\right],
\label{eq:trilinear-value}
\end{equation}
where $\mathcal N(i)$ contains the child cells connected to parent cell $i$ and $\gamma$ is a learned per-channel LayerScale initialized to one. The relation state starts from $r_\rho=\mathrm{RMSNorm}(W_{\mathrm{rel}}t_\rho)$, where $t_\rho$ is a frozen text embedding of the foreign-key description. We keep $r_\rho$ static across the 12 blocks, but $W_{\mathrm{rel}}$, the relation key/value projections, and LayerScale remain trainable. A sparse softmax evaluates only matched foreign-key edges, so memory scales with the number of typed edges.

\paragraph{Muon.} Adam rescales each coordinate by its own gradient statistics, so the update matrix concentrates on a few dominant directions and successive steps reinforce the same subspace. Muon replaces the momentum update of each matrix-valued parameter with the nearest semi-orthogonal matrix, computed by a five-step Newton--Schulz iteration in bfloat16 \citep{jordan2024muon}, so every singular direction advances comparably and a step buys more of the representation than an Adam step of the same norm. The gain is largest early in training, the only regime a speedrun measures, and RT is almost entirely matrix-valued, with four attention sublayers and a gated MLP per block. We apply it to the two-dimensional matrices in the standard attention, typed attention, and feed-forward modules, at peak learning rate $5\times10^{-3}$. AdamW updates the encoders, decoders, normalization and scale parameters, biases, and the SIGReg projector, at learning rate $10^{-3}$ and weight decay $0.1$. Both branches follow the reference schedule. Muon is inherited from prior speedruns and from relational pretraining \citep{liu2025muon,ranjan2026largescale}; we report its contribution but do not claim it.

\paragraph{SIGReg.} RT admits every column of every database into one residual stream through per-datatype encoders, so a price and a free-text title arrive with no shared scale or shape, and nothing prevents a column's hidden states from collapsing onto a low-dimensional region early in training; the shared blocks then spend capacity renormalizing their inputs. An isotropy constraint cannot add capacity, so we expect it to shorten the path to a given quality without raising the ceiling, which is what \citet{merchant2026optimizing} report for tabular models and what \citet{akbar2026weak} report for vision and MLP backbones.

SIGReg matches random one-dimensional projections of a representation distribution to a standard Gaussian through an Epps--Pulley statistic \citep{balestriero2025lejepa}. We apply it to the hidden states leaving the first block, so that the constraint acts on the backbone's input. Cells are grouped by (table, column), which enforces isotropy within each column's own population; a trainable $256\rightarrow1024\rightarrow64$ projector maps each hidden state, $K=64$ random unit vectors slice the output, and for every group with at least eight valid cells we penalize the squared distance between the empirical characteristic function of each slice and that of $\mathcal N(0,1)$ over $[-3,3]$. Writing that group-weighted penalty as $\mathcal R_{\mathrm{col}}$, the objective is
\begin{equation}
\mathcal L=(1-\lambda)\mathcal L_{\mathrm{pred}}+\lambda\mathcal
R_{\mathrm{col}},\qquad \lambda=10^{-3},
\label{eq:sigreg-objective}
\end{equation}
and the projector trains jointly with the backbone.

\begin{figure}[t]
\centering
\resizebox{\linewidth}{!}{%
\begin{tikzpicture}[
    x=1cm, y=1cm, font=\sffamily, >=Latex,
    flow/.style     ={-{Latex[length=1.9mm,width=1.3mm]}, draw=archmuted,
                      line width=0.75pt},
    newflow/.style  ={-{Latex[length=1.9mm,width=1.3mm]}, draw=archburgundy,
                      line width=1.0pt, line cap=round, line join=round},
    trainflow/.style={-{Latex[length=1.9mm,width=1.3mm]}, draw=archburgundy,
                      line width=1.0pt, dash pattern=on 2pt off 1.5pt,
                      line cap=round},
    box/.style      ={draw=archslate, fill=archslatefill, rounded corners=1.0mm,
                      line width=0.7pt, minimum width=14mm, minimum height=10mm,
                      align=center, inner sep=0.4mm, text width=13mm,
                      font=\scriptsize, text=archink},
    newbox/.style   ={box, draw=archburgundy, fill=archburgundyfill,
                      line width=1.2pt, text=archburgundy,
                      font=\scriptsize\bfseries},
    ghost/.style    ={draw=none, fill=none, minimum width=1.6mm,
                      minimum height=11mm, inner sep=0pt},
    shell/.style    ={draw=archline, fill=archshell, rounded corners=1.2mm,
                      line width=0.7pt, align=center, font=\scriptsize,
                      text=archink, inner sep=1mm},
    newshell/.style ={shell, draw=archburgundy, fill=archburgundyfill,
                      line width=1.0pt, text=archburgundy},
    note/.style     ={font=\scriptsize, text=archmuted, align=center},
]

\foreach \cy in {0.44,-0.44}{
  \path[draw=archline, fill=white, line width=0.6pt, rounded corners=0.4mm]
      (0.16,\cy-0.27) rectangle (1.04,\cy+0.27);
  \path[fill=archslatefill] (0.16,\cy+0.12) rectangle (1.04,\cy+0.27);
  \draw[archline, line width=0.4pt] (0.16,\cy+0.12) -- (1.04,\cy+0.12);
  \draw[archline, line width=0.4pt] (0.16,\cy-0.08) -- (1.04,\cy-0.08);
  \draw[archline, line width=0.4pt] (0.60,\cy-0.27) -- (0.60,\cy+0.12);
}
\draw[{Latex[length=1.4mm,width=1mm]}-{Latex[length=1.4mm,width=1mm]},
      draw=archmuted, line width=0.6pt] (0.60,0.15) -- (0.60,-0.15);
\node[note, text width=22mm] at (0.60,-1.02) {sampled rows,\\FK links};

\node[shell, text width=20mm] (enc) at (2.70,0)
    {\textbf{Per-type encoder}\\[1pt]
     {\fontsize{6}{7}\selectfont num $\cdot$ text $\cdot$ bool $\cdot$ time
      $+$ column name}};
\draw[flow] (1.14,0) -- (enc.west);

\draw[draw=archink!35, rounded corners=1.6mm, line width=0.85pt]
    (4.55,-0.85) rectangle (12.61,0.85);
\node[anchor=south west, font=\scriptsize\bfseries, text=archink]
    at (4.55,0.93) {RelationalBlock 1};

\node[box]    (col)  at (5.38,0) {Column\\attention};
\node[box]    (feat) at (6.98,0) {Feature /\\FK attention};
\node[newbox] (tri)  at (8.58,0) {Trilinear\\attention};
\node[box]    (full) at (10.18,0) {Full\\attention};
\node[box]    (ffn)  at (11.78,0) {SwiGLU\\MLP};

\foreach \a/\b in {col/feat,feat/tri,tri/full,full/ffn}{
  \draw[flow] (\a.east) -- (\b.west);
}
\draw[flow] (enc.east) -- (col.west);
\node[note, anchor=south] at (4.32,0.10) {$\mathbf x^{(0)}$};

\node[note, text width=22mm] (fkd) at (5.20,1.95)
    {foreign-key\\description text};
\node[newshell, text width=24mm] (renc) at (8.58,1.95)
    {\textbf{Relation encoder}\\[1pt]
     {\fontsize{6}{7}\selectfont FROZEN TEXT EMBEDDING}};
\draw[newflow] (fkd.east) -- (renc.west);
\draw[newflow] (renc.south) -- (tri.north);
\node[font=\scriptsize, text=archburgundy, anchor=west] at (8.68,1.05)
    {$\mathbf r_\rho$};

\coordinate (tap) at (12.88,0);
\node[circle, fill=archburgundy, inner sep=1pt] at (tap) {};
\node[newshell, text width=34mm] (sig) at (12.88,2.02)
    {\textbf{SIGReg}\\[1pt]
     {\fontsize{6}{7}\selectfont AFTER BLOCK 1 ONLY $\cdot$ TRAINING ONLY}};
\draw[trainflow] (tap) -- (sig.south);

\node[ghost] (g1) at (13.35,0) {};
\node[font=\large, text=archink!40] at (14.08,0) {$\cdots$};
\node[ghost] (g2) at (14.81,0) {};
\foreach \x in {13.35,14.81}{
  \path[draw=archburgundy, fill=archburgundyfill, line width=0.8pt,
        rounded corners=0.3mm] (\x-0.08,-0.55) rectangle (\x+0.08,0.55);
}
\draw[flow] (ffn.east) -- (g1.west);
\draw[flow] (g1.east) -- (13.83,0);
\draw[flow] (14.33,0) -- (g2.west);
\node[note, text width=24mm] at (14.08,-1.16) {blocks $2\ldots12$};

\node[shell, text width=15mm] (dec) at (16.01,0) {type decoder};
\node[draw=archline, fill=white, minimum width=4.8mm, minimum height=4.8mm,
      inner sep=0pt, font=\scriptsize, line width=0.6pt] (yh) at (17.31,0)
    {$\hat y$};
\draw[flow] (g2.east) -- (dec.west);
\draw[flow] (dec.east) -- (yh.west);
\node[note] at (16.01,-0.68) {predicted cell};

\end{tikzpicture}%
}
\caption{The two architectural changes in the record recipe, marked in burgundy. A frozen text embedding of each foreign-key description enters the trilinear operator, which replaces RT's neighbor attention in all twelve blocks. SIGReg is applied during training, at the first block only.}
\label{fig:architecture-nanort}
\end{figure}
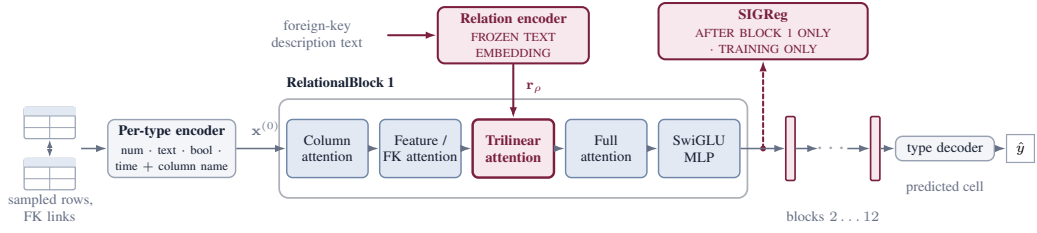

Figure~\ref{fig:architecture-nanort} summarizes the retained RT backbone and the three changes in the record recipe.

\section{Results}
\label{sec:results}

We report three views: the record lineage, a matched multi-seed sweep for attribution, and a leave-one-database-out study for transfer.

\subsection{The record}
\label{subsec:record}

\begin{figure}[H]
\centering

\begin{tikzpicture}
\begin{axis}[
    width=0.80\linewidth,
    y=4mm,
    xbar=9pt,
    bar width=9pt,
    symbolic y coords={+ Typed attention,+ SIGReg,+ Muon,Calibration horizon},
    ytick={+ Typed attention,+ SIGReg,+ Muon,Calibration horizon},
    yticklabels={+ Typed attention,+ SIGReg,+ Muon,Baseline Model},
    yticklabel style={font=\small,text=black!80},
    xmin=0,
    xmax=55,
    xtick={0,10,20,30,40,50},
    axis y line*=left,
    axis line style={draw=none},
    ytick style={draw=none},
    axis x line*=bottom,
    tick style={black!50},
    xmajorgrids=true,
    grid style={black!12},
    enlarge y limits={abs=4mm},
    clip=false,
]

\addplot[
    fill=controlgray!70,
    draw=none,
    bar shift=0pt,
]
coordinates {
    (41.3317,Calibration horizon) 
    (12.2700,+ Muon)
    (7.2633,+ SIGReg)
};

\addplot[
    fill=methodpurple,
    draw=none,
    bar shift=0pt,
] coordinates {(5.056,+ Typed attention)};

\node[anchor=west,font=\scriptsize,text=black!75]
    at (axis cs:41.3317,Calibration horizon)
at (axis cs:41.3317,Calibration horizon) {\;41:20};
\node[anchor=west,font=\scriptsize,text=black!75]
    at (axis cs:12.4500,+ Muon) {\;12:27 $\cdot$ $3.32\times$};
\node[anchor=west,font=\scriptsize,text=black!75]
    at (axis cs:7.4333,+ SIGReg) {\;7:26 $\cdot$ $5.56\times$};
\node[anchor=west,font=\scriptsize\bfseries,text=methodpurple]
    at (axis cs:5.0938,+ Typed attention) {\;5:05.6 $\cdot$ $8.11\times$};

\end{axis}
\node[anchor=north,font=\small]
    at ([yshift=-14pt]current axis.south)
    {Recorded duration (minutes) $\leftarrow$ shorter};
\end{tikzpicture}

\caption{
Cumulative record lineage, from the reference implementation's $41$:$20$ to $305.63$ seconds. Entries are chronological records, not controlled comparisons.
}
\label{fig:record-progression}
\end{figure}
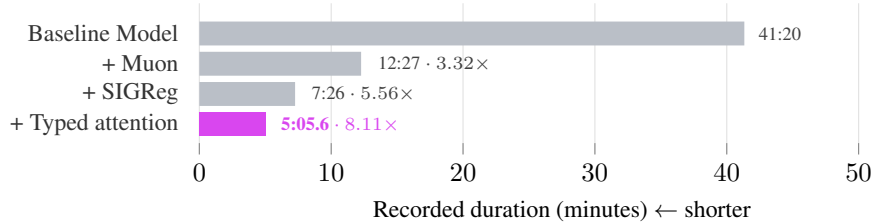

The archived modded-nanoRT record logs its first record-eligible verification at $305.63$ seconds ($5$:$05.6$) on one NVIDIA H200, $8.11\times$ faster than the reference implementation (Figure~\ref{fig:record-progression}).

That verification is the first scheduled evaluation, after 100 updates, and it returns $0.871392$ AUROC against a target of $0.81621$, above every value the reference implementation reaches inside its full 45-minute budget. The reference needs $10{,}500$ of its $11{,}621$ updates to reach the target and is evaluated every $500$, so both reported times are upper bounds set by their own grids. Verification dominates the measured duration, at $182.96$ seconds ($59.9\%$) against $122.67$ seconds of training ($40.1\%$). Those 100 updates also fall inside the learning-rate warmup, so the record is set before either optimizer reaches its peak rate; the schedules were chosen for the 45-minute horizon and are mis-specified for a run that now ends in five minutes. The benchmark now resolves time-to-target only to the width of its own evaluation grid, so the next entries should shorten the evaluation path, refine the evaluation schedule, and refit the learning-rate schedules before optimizing the update loop.

\subsection{Component attribution}

We evaluate six configurations with three paired seeds, scoring the best scheduled checkpoint within the budget. In Table~\ref{tab:component-contrasts}, Muon improves RT in all three pairs, by $+2.56$ points on average.

\begin{table}[h]
\centering
\caption{Paired change in best rel-Avito pooled AUROC (row minus control, percentage points). Mean $\pm$ standard deviation across three seeds; Wins counts seeds with a positive difference.}
\label{tab:component-contrasts}
\small
\setlength{\tabcolsep}{5pt}
\begin{tabular}{llrc}
\toprule
Component & Paired contrast & $\Delta$ AUROC (pp) & Wins \\
\midrule
\multirow{2}{*}{\emph{Muon}}
  & on AdamW RT            & $+2.56 \pm 3.13$ & 3/3 \\
  & with typed attention   & $+6.60 \pm 3.53$ & 3/3 \\
\multirow{2}{*}{\emph{Typed attention}}
  & on Muon$+$SIGReg       & $+3.56 \pm 4.79$ & 2/3 \\
  & on AdamW               & $-1.21 \pm 5.79$ & 1/3 \\
\multirow{2}{*}{\emph{SIGReg}}
  & before typed attention & $-0.47 \pm 0.66$ & 0/3 \\
  & with typed attention   & $+0.26 \pm 0.42$ & 3/3 \\
\bottomrule
\end{tabular}
\end{table}

Typed attention costs $1.21$ points under AdamW and wins one pair of three; under Muon$+$SIGReg it gains $3.56$ and wins two. Either the operator is simply slow, so a fixed budget measures convergence speed rather than compatibility, or the three-factor score needs orthogonalized updates that a coordinate-wise optimizer cannot supply.

Table~\ref{tab:component-contrasts} scores best AUROC within the budget, so it measures attainable quality, which Section~\ref{sec:recipe} predicts SIGReg should not move; both contrasts straddle zero ($-0.47$ and $+0.26$ points).  We therefore keep SIGReg in the record recipe on the time-to-target evidence of Section~\ref{subsec:record} and claim no effect for it on final quality.

\subsection{Transfer across held-out databases}

Across six leave-one-database-out holdouts, the final recipe wins 15 of 18 matched database--seed pairs, with mean and median gains of $2.14$ and $2.53$ AUROC points (Table~\ref{tab:relbench-best-paired}). Excluding development on rel-Avito, it wins 13 of 15 pairs and averages $+1.98$ points. Five databases have positive mean gains; rel-Amazon is the exception, and seed variance explains it. Best-checkpoint selection over three seeds ranks these contrasts; it does not test them.

\begin{table}[h]
\centering
\caption{Best pooled AUROC across leave-one-database-out holdouts. $\Delta$ is
Final minus AdamW RT in percentage points, mean $\pm$ standard deviation
across three seeds.}
\label{tab:relbench-best-paired}
\small
\setlength{\tabcolsep}{5pt}
\begin{tabular}{lrrrc}
\toprule
Held-out DB & AdamW RT & Final & $\Delta$ (pp) & Wins \\
\midrule
rel-Amazon & 58.45 & 58.17 & $-0.29 \pm 12.86$ & 2/3 \\
rel-Avito  & 80.21 & 83.15 & $+2.94 \pm \phantom{0}6.24$  & 2/3 \\
rel-F1     & 79.55 & 84.58 & $+5.04 \pm \phantom{0}1.69$  & 3/3 \\
rel-HM     & 62.77 & 65.14 & $+2.37 \pm \phantom{0}0.51$  & 3/3 \\
rel-Stack  & 75.07 & 76.96 & $+1.88 \pm \phantom{0}1.85$  & 3/3 \\
rel-Trial  & 51.50 & 52.38 & $+0.88 \pm \phantom{0}2.91$  & 2/3 \\
\midrule
All pairs  & 67.93 & 70.06 & $+2.14 \pm 5.36$  & 15/18 \\
\bottomrule
\end{tabular}
\end{table}

\section{Conclusion}
modded-nanoRT holds the problem fixed and leaves the training stack open, so that improvements accumulate across contributors. The opening entry reaches the target in $305.63$ seconds on one H200, $8.11\times$ faster than the reference implementation. Matched comparisons attribute that to Muon, and to typed attention only in combination with it, and outside the development database the recipe wins 13 of 15 pairs. Every reported quantity selects the best scheduled checkpoint within the budget over three seeds, and rel-Amazon's standard deviation of $12.86$ points exceeds every effect measured here. 

Verification consumes $59.9\%$ of the record and the first scheduled evaluation clears the target by $5.5$ points, so the next entry should attack the evaluation path before the update loop, refit the learning-rate schedules to the horizon the record now occupies, and drop the text encoder (Appendix~\ref{app:fk-ablation}).

\bibliographystyle{plainnat}
\bibliography{references}

@misc{robinson2024relbench,
  title         = {{RelBench}: A Benchmark for Deep Learning on Relational Databases},
  author        = {Joshua Robinson and Rishabh Ranjan and Weihua Hu and Kexin Huang and Jiaqi Han and Alejandro Dobles and Matthias Fey and Jan E. Lenssen and Yiwen Yuan and Zecheng Zhang and Xinwei He and Jure Leskovec},
  year          = {2024},
  eprint        = {2407.20060},
  archiveprefix = {arXiv},
  primaryclass  = {cs.LG},
  url           = {https://arxiv.org/abs/2407.20060}
}

@misc{ranjan2026relational,
  title         = {Relational Transformer: Toward Zero-Shot Foundation Models for Relational Data},
  author        = {Rishabh Ranjan and Valter Hudovernik and Mark Znidar and Charilaos Kanatsoulis and Roshan Upendra and Mahmoud Mohammadi and Joe Meyer and Tom Palczewski and Carlos Guestrin and Jure Leskovec},
  year          = {2026},
  note          = {Accepted at ICLR 2026},
  eprint        = {2510.06377},
  archiveprefix = {arXiv},
  primaryclass  = {cs.LG},
  url           = {https://arxiv.org/abs/2510.06377}
}

@misc{kothapalli2026plurel,
  title         = {{PluRel}: Synthetic Data Unlocks Scaling Laws for Relational Foundation Models},
  author        = {Vignesh Kothapalli and Rishabh Ranjan and Valter Hudovernik and Vijay Prakash Dwivedi and Johannes Hoffart and Carlos Guestrin and Jure Leskovec},
  year          = {2026},
  note          = {Accepted at ICML 2026},
  eprint        = {2602.04029},
  archiveprefix = {arXiv},
  primaryclass  = {cs.DB},
  url           = {https://arxiv.org/abs/2602.04029}
}

@misc{coleman2018dawnbench,
  title         = {Analysis of {DAWNBench}, a Time-to-Accuracy Machine Learning Performance Benchmark},
  author        = {Cody Coleman and Daniel Kang and Deepak Narayanan and Luigi Nardi and Tian Zhao and Jian Zhang and Peter Bailis and Kunle Olukotun and Chris R\'{e} and Matei Zaharia},
  year          = {2018},
  eprint        = {1806.01427},
  archiveprefix = {arXiv},
  primaryclass  = {cs.LG},
  url           = {https://arxiv.org/abs/1806.01427}
}

@misc{mattson2020mlperf,
  title         = {{MLPerf} Training Benchmark},
  author        = {Peter Mattson and others},
  year          = {2020},
  note          = {MLSys 2020},
  eprint        = {1910.01500},
  archiveprefix = {arXiv},
  primaryclass  = {cs.LG},
  url           = {https://arxiv.org/abs/1910.01500}
}

@misc{geiping2022cramming,
  title         = {Cramming: Training a Language Model on a Single {GPU} in One Day},
  author        = {Jonas Geiping and Tom Goldstein},
  year          = {2022},
  eprint        = {2212.14034},
  archiveprefix = {arXiv},
  primaryclass  = {cs.CL},
  url           = {https://arxiv.org/abs/2212.14034}
}

@misc{pfefferle2025nanotabpfn,
  title         = {{nanoTabPFN}: A Lightweight and Educational Reimplementation of {TabPFN}},
  author        = {Alexander Pfefferle and Johannes Hog and Lennart Purucker and Frank Hutter},
  year          = {2025},
  eprint        = {2511.03634},
  archiveprefix = {arXiv},
  primaryclass  = {cs.LG},
  url           = {https://arxiv.org/abs/2511.03634}
}

@misc{ozturk2026speedrunning,
  title         = {Speedrunning Tabular Foundation Model Pretraining},
  author        = {Salih Bora Ozturk and Alexander Pfefferle and Frank Hutter},
  year          = {2026},
  eprint        = {2606.03681},
  archiveprefix = {arXiv},
  primaryclass  = {cs.LG},
  url           = {https://arxiv.org/abs/2606.03681},
  note = {2nd ICML Workshop on Foundation Models for Structured Data}
}

@misc{jordan2024muon,
  title  = {{Muon}: An Optimizer for Hidden Layers in Neural Networks},
  author = {Keller Jordan and Yuchen Jin and Vlado Boza and {You Jiacheng} and Franz Cesista and Laker Newhouse and Jeremy Bernstein},
  year   = {2024},
  url    = {https://kellerjordan.github.io/posts/muon/}
}

@misc{liu2025muon,
  title         = {{Muon} Is Scalable for {LLM} Training},
  author        = {Jingyuan Liu and Jianlin Su and Xingcheng Yao and Zhejun Jiang and Guokun Lai and Yulun Du and Yidao Qin and Weixin Xu and Enzhe Lu and Junjie Yan and Yanru Chen and Huabin Zheng and Yibo Liu and Shaowei Liu and Bohong Yin and Weiran He and Han Zhu and Yuzhi Wang and Jianzhou Wang and Mengnan Dong and Zheng Zhang and Yongsheng Kang and Hao Zhang and Xinran Xu and Yutao Zhang and Yuxin Wu and Xinyu Zhou and Zhilin Yang},
  year          = {2025},
  eprint        = {2502.16982},
  archiveprefix = {arXiv},
  primaryclass  = {cs.LG},
  url           = {https://arxiv.org/abs/2502.16982}
}

@misc{balestriero2025lejepa,
  title         = {{LeJEPA}: Provable and Scalable Self-Supervised Learning Without the Heuristics},
  author        = {Randall Balestriero and Yann LeCun},
  year          = {2025},
  eprint        = {2511.08544},
  archiveprefix = {arXiv},
  primaryclass  = {cs.LG},
  url           = {https://arxiv.org/abs/2511.08544}
}

@misc{schlichtkrull2017rgcn,
  title         = {Modeling Relational Data with Graph Convolutional Networks},
  author        = {Michael Schlichtkrull and Thomas N. Kipf and Peter Bloem and Rianne van den Berg and Ivan Titov and Max Welling},
  year          = {2017},
  eprint        = {1703.06103},
  archiveprefix = {arXiv},
  primaryclass  = {stat.ML},
  url           = {https://arxiv.org/abs/1703.06103}
}

@misc{clift2019logic,
  title         = {Logic and the $2$-Simplicial Transformer},
  author        = {James Clift and Dmitry Doryn and Daniel Murfet and James Wallbridge},
  year          = {2019},
  eprint        = {1909.00668},
  archiveprefix = {arXiv},
  primaryclass  = {cs.LG},
  url           = {https://arxiv.org/abs/1909.00668}
}

@misc{roy2025fast,
  title         = {Fast and Simplex: $2$-Simplicial Attention in {Triton}},
  author        = {Aurko Roy and Timothy Chou and Sai Surya Duvvuri and Sijia Chen and Jiecao Yu and Xiaodong Wang and Manzil Zaheer and Rohan Anil},
  year          = {2025},
  eprint        = {2507.02754},
  archiveprefix = {arXiv},
  primaryclass  = {cs.LG},
  url           = {https://arxiv.org/abs/2507.02754}
}

@misc{modded_nanogpt_2024,
  author       = {Keller Jordan and Jeremy Bernstein and Brendan Rappazzo and
                  @fernbear.bsky.social and Boza Vlado and You Jiacheng and
                  Franz Cesista and Braden Koszarsky and @Grad62304977},
  title        = {modded-nanogpt: Speedrunning the NanoGPT baseline},
  year         = {2024},
  url          = {https://github.com/KellerJordan/modded-nanogpt}
}

@inproceedings{
merchant2026optimizing,
title={Optimizing Pre-Training of Tabular Foundation Models by Shaping Geometry},
author={Humzah Merchant and Sriniketh Vangaru and Randall Balestriero},
booktitle={2nd ICML Workshop on Foundation Models for Structured Data},
year={2026},
url={https://openreview.net/forum?id=IYnHchzvYB}
}

@inproceedings{
ranjan2026largescale,
title={Large-Scale Pretraining unlocks Few-Shot Prediction for Relational Data},
author={Rishabh Ranjan and Vignesh Kothapalli and Harshvardhan Agarwal and Charilaos I. Kanatsoulis and Roshan Reddy Upendra and Tom Palczewski and Carlos Guestrin and Jure Leskovec},
booktitle={2nd ICML Workshop on Foundation Models for Structured Data},
year={2026},
url={https://openreview.net/forum?id=oQINTd9din}
}

@inproceedings{xue2021mt5,
  title     = {{mT5}: A Massively Multilingual Pre-trained Text-to-Text Transformer},
  author    = {Xue, Linting and Constant, Noah and Roberts, Adam and Kale, Mihir
               and Al-Rfou, Rami and Siddhant, Aditya and Barua, Aditya and Raffel, Colin},
  booktitle = {NAACL-HLT},
  pages     = {483--498},
  year      = {2021}
}

@article{akbar2026weak,
  title={Weak-SIGReg: Covariance Regularization for Stable Deep Learning},
  author={Akbar, Habibullah},
  journal={arXiv preprint arXiv:2603.05924},
  year={2026}
}

\appendix

\section{What the relation vector must carry}
\label{app:fk-ablation}

The third attention factor uses a frozen embedding of each foreign-key description, which may contribute semantic content or simply distinguish relation types. Four conditions isolate these effects over seven seeds, with the rest of the recipe fixed (Table~\ref{tab:fk-ablation}). \emph{Semantic FK} is the record recipe; \emph{Shuffled FK} permutes its embeddings across types; \emph{Random per type/layer} assigns each type a fixed random vector in each block; and \emph{Random per edge} assigns one to each edge, removing type identity.

\begin{table}[h]
\centering
\caption{Relation-vector ablation on rel-Avito over seven seeds (mean $\pm$ standard deviation). Reached target counts seeds whose best pooled AUROC exceeds $0.81621$.}
\label{tab:fk-ablation}
\small
\setlength{\tabcolsep}{6pt}
\begin{tabular}{lccc}
\toprule
Relation vector & Best AUROC & Steps 100--900 & Reached target \\
\midrule
Semantic FK (record recipe) & $0.8335 \pm 0.0330$ & $0.7656 \pm 0.0327$ & 5/7 \\
Shuffled FK & $0.8371 \pm 0.0381$ & $0.7794 \pm 0.0395$ & 5/7 \\
Random per type/layer & $\mathbf{0.8397} \pm 0.0156$ & $\mathbf{0.7830} \pm 0.0254$ & \textbf{7/7} \\
Random per edge & $0.8204 \pm 0.0281$ & $0.7739 \pm 0.0328$ & 3/7 \\
\bottomrule
\end{tabular}
\end{table}

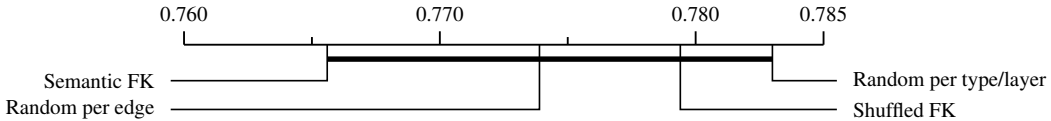
\begin{figure}[h]
\centering
\noindent\resizebox{\linewidth}{!}{%
\begin{tikzpicture}[
    x=1cm,
    y=0.72cm, 
    every node/.style={font=\small},
    method/.style={font=\small},
    axisline/.style={draw=black, line width=0.8pt},
    connector/.style={draw=black, line width=0.7pt},
    clique/.style={draw=black, line width=2.3pt, line cap=butt}
]

\pgfmathsetmacro{\semanticx}{(0.7656-0.760)/(0.785-0.760)*9.6}
\pgfmathsetmacro{\edgex}{(0.7739-0.760)/(0.785-0.760)*9.6}
\pgfmathsetmacro{\shufflex}{(0.7794-0.760)/(0.785-0.760)*9.6}
\pgfmathsetmacro{\typex}{(0.7830-0.760)/(0.785-0.760)*9.6}

\draw[axisline] (0,0) -- (9.6,0);
\foreach \x/\label in {
    0/0.760,
    3.84/0.770,
    7.68/0.780,
    9.6/0.785
} {
  \draw[axisline] (\x,0) -- (\x,0.28);
  \node[anchor=south] at (\x,0.31) {\label};
}
\foreach \x in {1.92,5.76} {
  \draw[axisline] (\x,0) -- (\x,0.16);
}

\draw[clique] (\semanticx,-0.30) -- (\typex,-0.30);

  \draw[connector]
    (\semanticx,0) -- (\semanticx,-0.75) -- (-0.20,-0.75);
  \node[method,anchor=east] at (-0.32,-0.75) {Semantic FK};

  \draw[connector]
    (\edgex,0) -- (\edgex,-1.35) -- (-0.20,-1.35);
  \node[method,anchor=east] at (-0.32,-1.35) {Random per edge};

  \draw[connector]
  (\typex,0) -- (\typex,-0.75) -- (9.80,-0.75);
\node[method,anchor=west] at (9.92,-0.75)
  {Random per type/layer};

\draw[connector]
  (\shufflex,0) -- (\shufflex,-1.35) -- (9.80,-1.35);
\node[method,anchor=west] at (9.92,-1.35)
  {Shuffled FK};
\end{tikzpicture}%
}
\caption{Mean AUROC over steps 100--900. The bar denotes the nonsignificance clique at $\alpha=0.05$.}
\label{fig:fk-cd}
\end{figure}

Shuffling does not reduce either reported mean, while fixed random type/layer vectors give the highest means, reach the target in every seed, and have less than half the record recipe's best-AUROC standard deviation. Removing type identity (\emph{Random per edge}) reaches the target in only three seeds. However, no pair separates at $\alpha=0.05$ ($\min p_{\mathrm{Holm}}=0.441$; Figure~\ref{fig:fk-cd}), so this ordering is descriptive. The results support dropping the text encoder in later entries.

\section{Reference calibration trajectory}
\label{app:calibration}

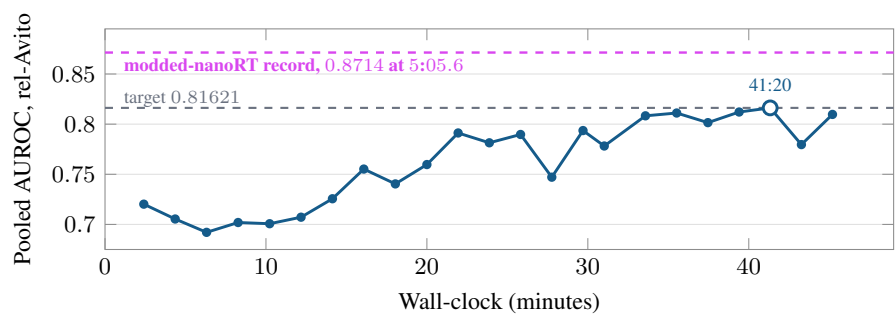
\begin{figure}[h]
\centering
\begin{tikzpicture}
\begin{axis}[
    width=0.86\linewidth, height=4.5cm,
    xlabel={Wall-clock (minutes)},
    ylabel={Pooled AUROC, rel-Avito},
    xmin=0, xmax=49, ymin=0.675, ymax=0.895,
    xtick={0,10,20,30,40}, ytick={0.70,0.75,0.80,0.85},
    yticklabel style={/pgf/number format/fixed,/pgf/number format/precision=2},
    axis line style={draw=black!45}, tick style={black!45},
    ymajorgrids=true, grid style={black!12},
    label style={font=\small}, tick label style={font=\small},
    clip=false,
]
\addplot[targetgray, dashed, line width=0.8pt, forget plot]
    coordinates {(0,0.81621) (49,0.81621)};
\node[anchor=south west, font=\scriptsize, text=targetgray] at (axis cs:0.6,0.8075) {target $0.81621$};

\addplot[nanortblue, line width=1.1pt, mark=*, mark size=1.2pt,
         mark options={fill=nanortblue,draw=none}]
  coordinates {
    (2.41,0.7202) (4.36,0.7054) (6.31,0.6921) (8.27,0.7019) (10.23,0.7007)
    (12.18,0.7072) (14.13,0.7256) (16.08,0.7552) (18.04,0.7404) (20.00,0.7597)
    (21.94,0.7912) (23.88,0.7814) (25.82,0.7897) (27.76,0.7471) (29.70,0.7936)
    (31.02,0.7782) (33.58,0.8083) (35.52,0.8111) (37.46,0.8015) (39.40,0.8121)
    (41.33,0.8162) (43.27,0.7796) (45.20,0.8097)
  };

\addplot[only marks, mark=*, mark size=2.6pt,
         mark options={fill=white, draw=nanortblue, line width=1.1pt}]
    coordinates {(41.33,0.8162)};
\node[anchor=south, font=\scriptsize, text=nanortblue] at (axis cs:41.33,0.8235) {41:20};

\addplot[methodpurple, dashed, line width=0.9pt, forget plot]
    coordinates {(0,0.8714) (49,0.8714)};
\node[anchor=north west, font=\scriptsize\bfseries, text=methodpurple] at (axis cs:0.6,0.875) {modded-nanoRT record, $0.8714$ at $5$:$05.6$};
\end{axis}
\end{tikzpicture}
\caption{Reference evaluation trajectory over the 45-minute calibration budget ($11{,}621$ updates, evaluated every $500$). Its best value, $0.81621$ at $41$:$20$, defines the target; the record recipe exceeds it at $5$:$05.6$, reaching $0.8714$. Times are upper bounds determined by the evaluation grids.}
\label{fig:calibration-curve}
\end{figure}

\end{document}